\documentclass{article} 

\usepackage{iclr2020_conference,times}

\usepackage{comment}
\usepackage{hyperref}
\usepackage{url}
\usepackage{graphicx}
\usepackage{subfig}
\usepackage{float}

\title{Estimating Grape Yield on the Vine \\ from Multiple Images}

\author{Daniel L. Silver and Jabun Nasa \\
Jodrey School of Computer Science, Acadia University, Wolfville, NS 15213,Canada \\
\texttt{\footnotesize danny.silver@acadiau.ca, 082392n@acadiau.ca} \\
}

\iclrfinalcopy 
\begin{document}

\maketitle

\begin{footnotesize}
\begin{abstract}
Estimating grape yield prior to harvest is important to commercial vineyard production as it
informs many vineyard and winery decisions.
Currently, the process of yield estimation is time consuming and varies in its accuracy from 75-90\% depending on the experience of the viticulturist. This paper proposes a multiple task learning (MTL) convolutional neural network (CNN) approach that uses images captured by inexpensive smart phones secured in a simple tripod arrangement. The CNN models use MTL transfer from autoencoders to achieve 85\% accuracy from image data captured 6 days prior to harvest.
\end{abstract}
\end{footnotesize}

\section{Introduction}
For viticulturists and vinters, accurate crop yield estimates are essential for planning the harvest, estimating storage requirements, making wine production choices and managing human and material resources in the field and the winery \citep{komm2015vineyard}.
However, making an accurate yield estimation (weight in grams) is a challenging and time consuming process.
Viticulturists traditionally estimate grape yield using a three step process.
They sample several clusters  of grapes from a block of the vineyard, weigh each cluster and compute the average cluster weight $(cw)$.  They then count the number of clusters on several vines and compute the average number of clusters per vine $(cv)$.  They then estimate the total yield $= cw \times cv  \times$ number of vines in the block.   
Viticulturists will err in their estimates from 10-25\% depending upon their experience.

There is value in having better methods based on data that is captured from inexpensive technology in the field, such as the combination of smartphones and cloud-based computing. 
We investigate methods of estimating the weight (in grams) of grapes on the vine using low-cost smartphone images prior to harvest.  
Averaging the weight from a number of images, would allow the vineyard to better estimate the yield for a block. 
This is an interesting  problem because the predictive model must  estimate the weight of the grapes when many are occluded by other grapes.  
We know this is possible because viticulturists are able to make reasonable estimates based on having seen many harvests.  

In the past, various grape image segmentation techniques such as threshold segmentation, Mahalanobis distance segmentation, Bayesian classifier, direct three-dimensional histogram and linear color models have been investigated \citep{font2015vineyard}.  
Grape berry detectors based on support vector machines (SVM) and histograms of oriented gradients (HOG) features have proven to be efficient in the detection of white grape varieties \citep{vskrabanek2016evaluation}. 
And the weight of picked grape clusters lit under ideal conditions has been predicted using numerous features such as pixel area, volume, perimeter, berry number and berry size  \citep{nuske2011yield,diago2012grapevine}. 
  
We have previously presented a solution to this problem using a hand held smart phone \citep{SilverMonga2019}.  
Using a transfer learning approach that requires labelling the location of the grapes in the training images, we obtained a test mean absolute error (MAE) of 157.32 grams per half vine, or a mean accuracy if 88.21\%.  This is a great result, however  the data used in developing and testing these models was captured on the day of harvest and required a calibration marker carefully placed in each image proximal to the grapes.

We seek an approach that is equally accurate but does not require marked-up images taken with a calibration marker nor on the day of harvest.  
In this paper we present an approach that is able to estimate Pinot Noir grape yield up to 16 days in advance of harvest to 82\% mean accuracy on independent test sets of images.  
The approach is motivated by the process that humans take when estimating a metric such as volume or weight - they scan the object from as many angles as possible.  
We show that (1) more accurate models for estimating crop yield on the vine can be developed using deep convolution neural network (CNN) methods that accept multiple images taken by an inexpensive smart phone configuration, and that (2) multiple task learning (MTL) models that develop autoencoders of these multiple images can be used to improve the yield accuracy without the need for calibration markers or marked-up images
\citep{Caruana97}.  


\section{Approach}
\label{approach}

\begin{figure}[t]
    \centering
    \subfloat[Dual camera and tripod.]{
	\includegraphics[height=1.3in,keepaspectratio]{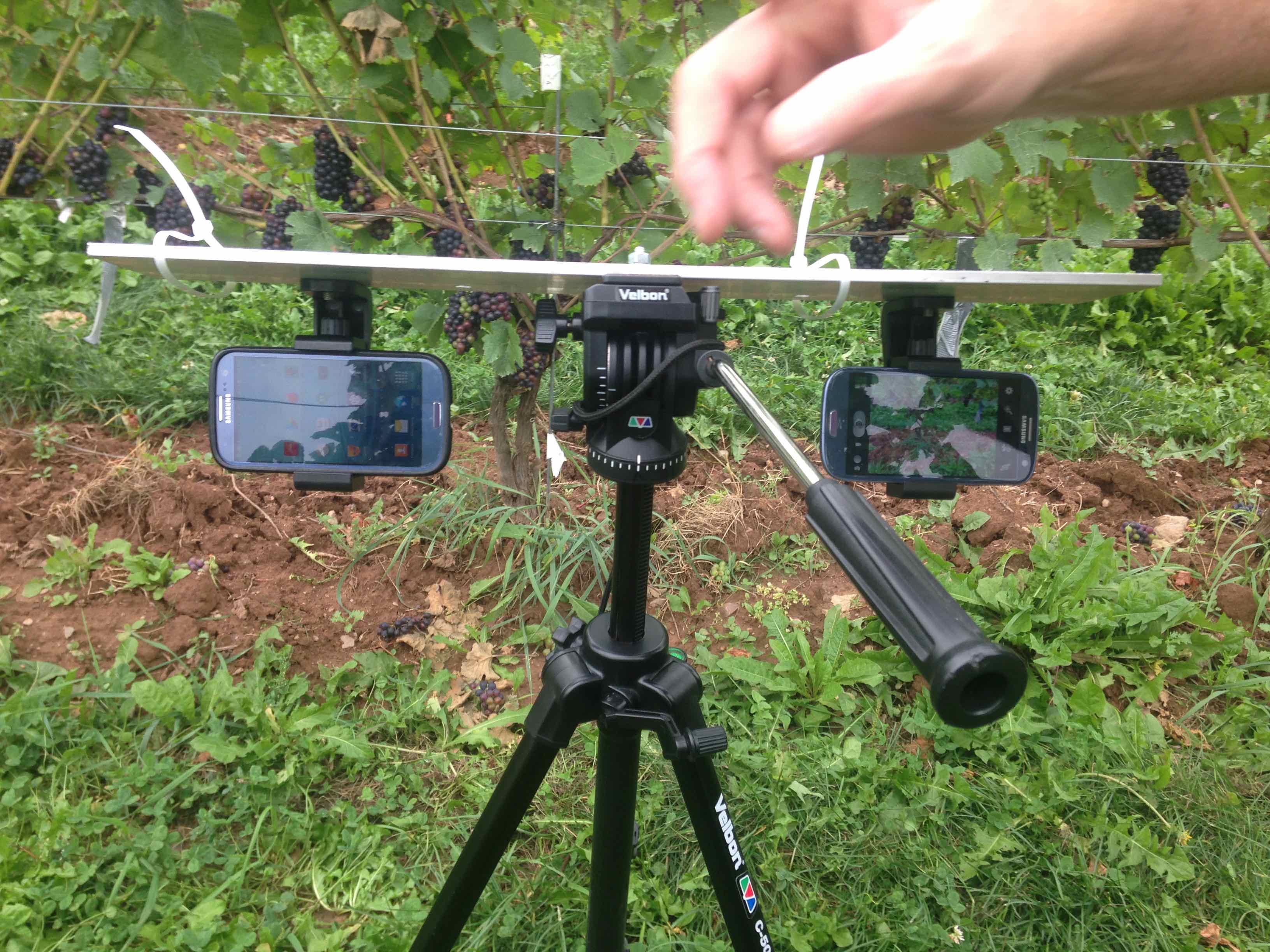}
    }
     \subfloat[Camera configuration and cropped images.]{
	\includegraphics[width=2.8in,height=1.3in]{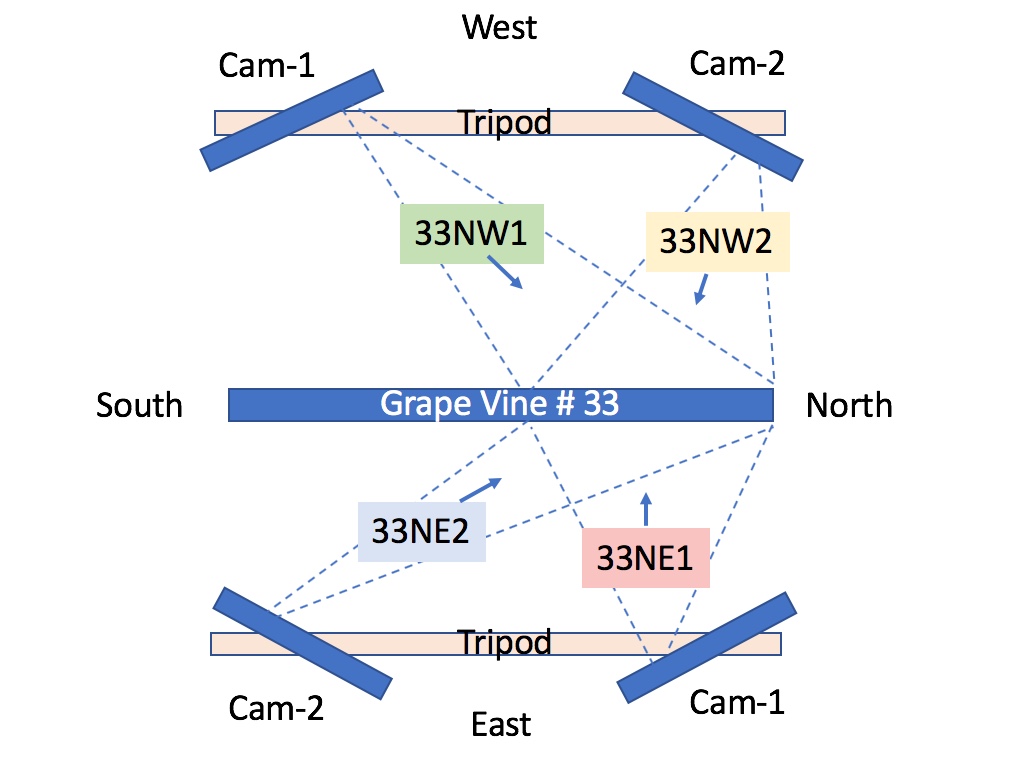}
    }
 \caption{Camera tripod configuration and the associated names for the four cropped images for a vine half, also called a vine cordon. The identifier of each image (\emph{e.g.} 33NE1) contains the plant number (33), vine cordon (N), vine side (E), and camera number (1).}
\label{camera_tripod}
\end{figure}

\paragraph{Image capture and yield collection.}
The data used in this research are images of Pinot Noir grapes on the vine and their associated harvest weight in grams per image.  The images were captured at the Lightfoot \& Wolfville Vineyards of Wolfville, Nova Scotia 6 days and 16 days prior to harvest, on October 23, 2018. 
The cameras on two Samsung Galaxy S3 smartphones (labelled Cam-1 and Cam-2) were used at a resolution of 3264$\times$2448 pixels.  The smartphones were mounted at either ends of a metal bar that was set in a fixed height tripod and angled to capture both sides of each vine (see Figure~\ref{camera_tripod}). Two cords tied to the tripod were used to position the cameras  the same distance from, in the centre of, and parallel to the vine (+/- a few inches).  No calibration marker was used. Each plant was numbered and pictures were taken of both sides of every plant using the dual camera configuration.
Four photographs of 80 vines were taken for a total of 320 original images in total.
Each image was cropped to contain just the grapes on one half of a vine, called a cordon.  
The north cordon of the vine is designated N and the south cordon is designated S.
The east-facing side of the vine is designated E and the west-facing side of the vine is designated  W.
The image of the grapes on the east side of the north cordon of plant 33 taken by camera 1 was recorded as image 33NE1 and the grapes on the west side of the south cordon of the plant 42 taken by camera 2 was labelled 42SW2. 
Each cordon of each vine was individually picked and weighed allowing us to create 160 examples, each composed of a set of four images and its associated weight for a cordon.   There was on average of about 1200 grams per cordon, vary depending upon the selection of cordons.

\paragraph{Image preprocessing.}
Cropping an image to capture just the grapes on a cordon produces images of different sizes. Resizing these images to a common size for the neural network is necessary but it would change the scale of the grapes from their original size. To maintain the original scale of the images white padding was added to each cropped image to make it the same size and only then were the images resized to 150 x 150 x 3 pixels required by the neural network.


Differences in the direction of the sunlight and changes in cloud cover required the images to be normalized as much as possible. Histogram equalization was used to correct the brightness of each image and then histogram matching was applied to normalize each to a standard image level of colour and brightness \citep{Matthew:2016}.

\paragraph{Multi Task Learning (MTL).}

\begin{figure}[h]
\centering
\includegraphics[width=4.2in, height=2.5in]{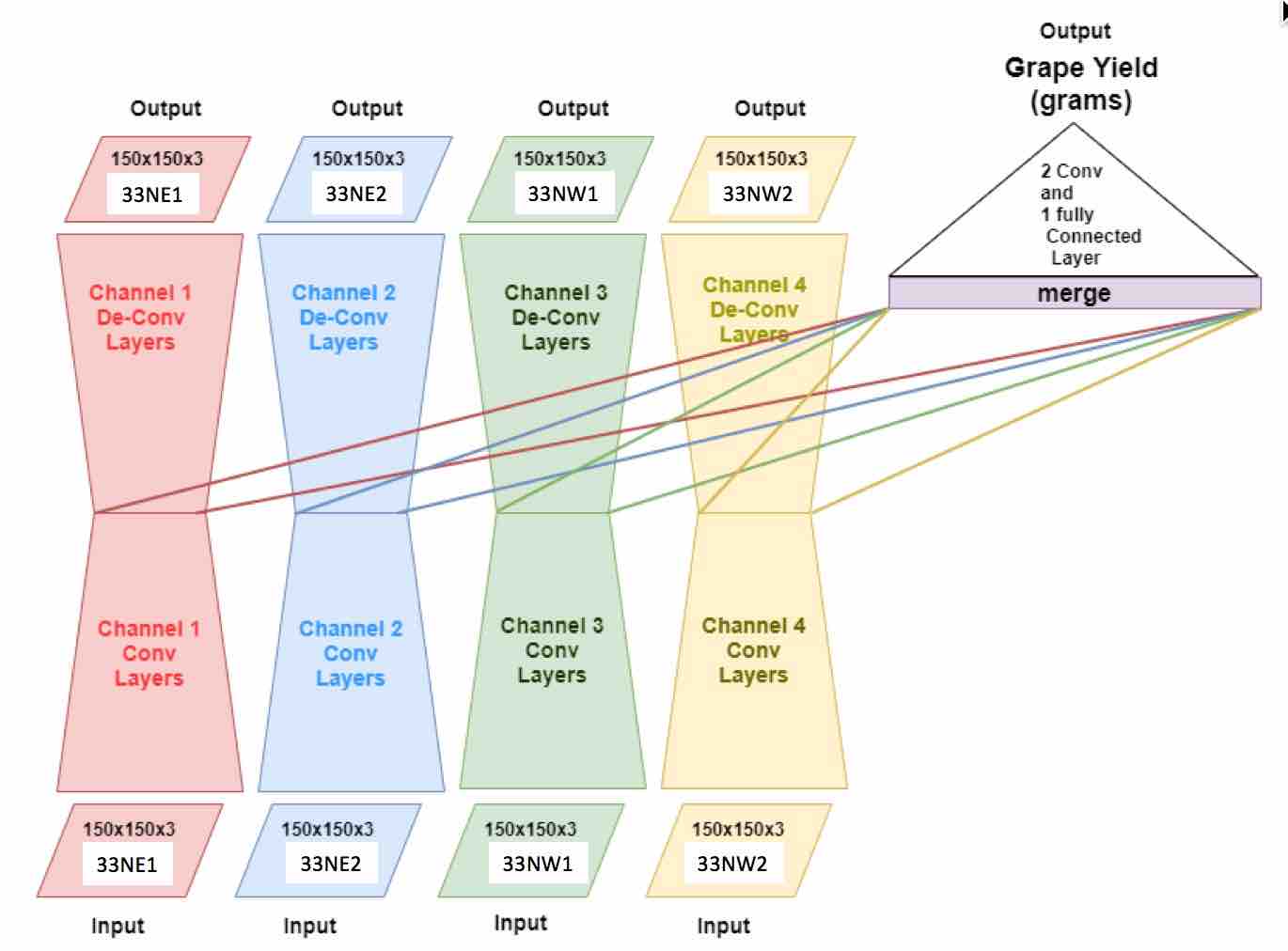}
\caption{Four image input CNN MTL network with five outputs (4 autoencoder,1 weight output)}
\label{CNN_MTL_network}
\end{figure}

Tensorflow and keras libraries routines are used to develop the deep MTL CNN network shown in Figure~\ref{CNN_MTL_network}. The images are cropped, resized and preprocessed 150 x 150 RGB pixels to feed the MTL. The MTL model takes 4 images of the same cordon as an input and produces 4 images as output as well as a grape yield estimation (in grams). 
All of the four input channels have four convolutional layers of the same size and each independently works with its own output channel composed of five deconvolution layers to produce an autoencoder for its input.  The autoencoder features created at the top convolution layer are merged and used as input to a yield estimation channel that consists of 2 convolutional layers and 1 fully connected block.  The model is trained using a validation set for the grape yield output to prevent over-fitting the model.   Note that this approach provides the potential for using additional unlabeled image data to develop more accurate autoencoders that can benefit MTL development of the crop yield model.

\section{Model Development and Testing}

\paragraph{Objective.}

The objectives of these experiments are: (1) To compare the yield estimation performance of a Single Task Learning (STL) network that accepts images taken from 4 different angles per cordon with a STL model that accepts images taken from only one angle; and (2) To investigate the performance of a 4-channel Multiple Task Learning (MTL) model developed from images collected up to 16 days prior to harvest.   The data and neural network used is as described in Section~\ref{approach}.     

\paragraph{Experiment 1: Single Task Learning using 1 versus 4 Images of Grapes.}

\begin{figure}[t]
\centering
\includegraphics[width=3.6in, height=1.2in]{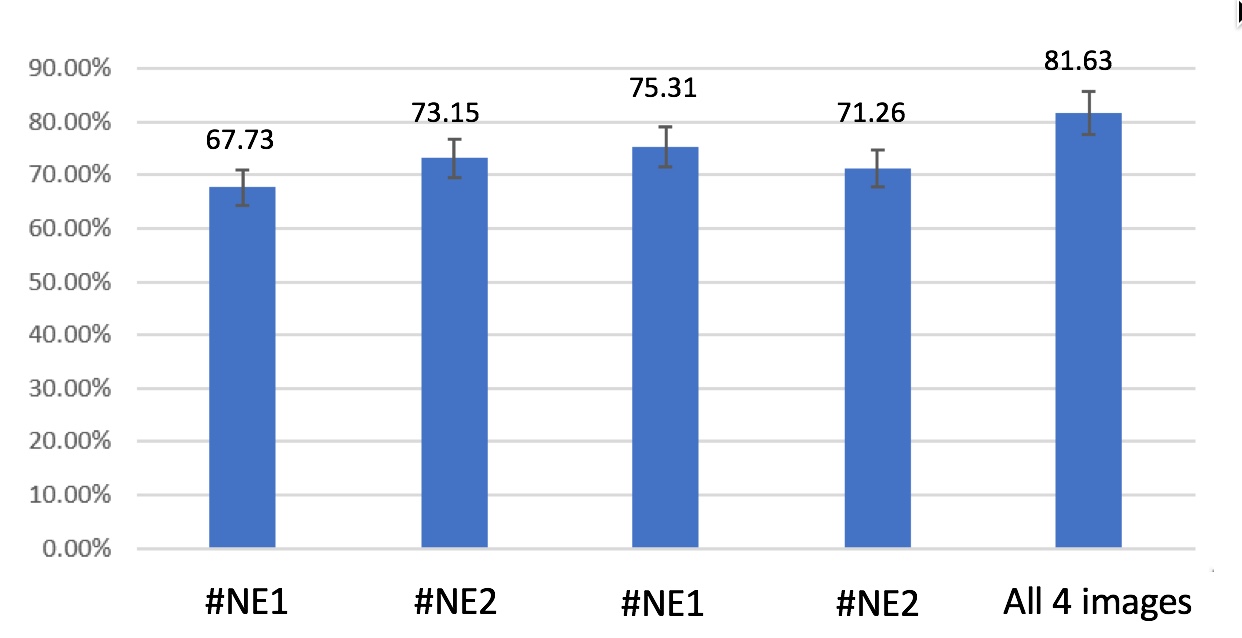}
\caption{Comparison of CNN Single Task Learning models. \#EN = images from east side north cordon, \#ES = east side south cordon, \#WS = west side south cordon, \#WN = west side north cordon.}
\label{model_results}
\end{figure}

An STL model was developed and tested using all north cordon, east side images from camera 1 (\emph{e.g.} 33NE1) as input to all channels of the network described in Section~\ref{approach}.  The autoencoder output channels were not used for this experiment.  A 6-fold cross-validation approach was used.  This process was repeated, each time using only the images from one of the other three image angles. 
Finally, the images from all four angles for each cordon were input, each to a channel, to train the same network architecture to predict the grape yield for that cordon.  
All images were taken 6 days prior to harvest, making this a more challenging problem than described in \citep{SilverMonga2019}.
The mean absolute error, or MAE, in grams and mean accuracy = 1 - (MAE / mean grams per cordon) are used as performance measures. 
The results shown in Figure~\ref{model_results} indicate that there is significant value in using multiple images from around the grape vine to estimate grape yield.  The 4-image CNN models were able to predict grape yield with an MAE of 231.51 grams or 81.63\% accuracy.

\paragraph{Experiment 2: Multiple Task Learning using 4 Images of Grapes.}

\begin{figure}[t]
    \centering
    \subfloat[Original image.]{
	\includegraphics[height=1.1in,keepaspectratio]{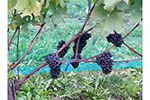}
    }
     \subfloat[Reconstructed image.]{
	\includegraphics[height=1.1in,keepaspectratio]{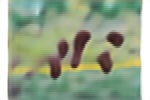}
    }
 \caption{Original image vs reconstructed images from best MTL model.}
\label{original_reconstructed}
\end{figure}

An MTL model was developed and tested using the network described in Section~\ref{approach} and images taken from all four angles of the grapes on each cordon, 6 days prior to harvest.  This time the autoencoder output channels were trained in parallel as a source of inductive bias for the grape yield regression output channel.  A 6-fold cross-validation approach was used.  A second model was developed and tested using data from 16 days prior to harvest using the same approach. The same performance metrics used in Exp 1 were used for this experiment. 
The MAE of the MTL models using data capture 6 days prior to harvest was 166.35 grams (95\% CI = 7.84) or 85.15\% accuracy.  This is significantly better than the best STL model with  97\% confidence. The autoencoders reconstructed remarkably accurate test images with an mean squared error per pixel of 0.0206 (see Figure~\ref{original_reconstructed}).    Even more surprising were the results using image data from 16 days prior to harvest.  The MAE of MTL CNN models developed from this data was 216.20 grams or 82\% accuracy;  a loss of only 50 grams or about 3\% accuracy.

\section{Conclusion and Future work}

The empirical studies show that the proposed MTL model achieved 82\% mean accuracy when trained and tested with images 16 days prior to harvest and 85\% accuracy using images 6 days prior to harvest.  This is within the range of accuracy achieved by viticulturists with considerable experience.  This is quite remarkable given our previously reported results of 88.21\% accuracy \citep{SilverMonga2019}. These models used images acquired the day of the harvest, preprocessed with the aid of a calibration marker, and took advantage of transfer learning from models that produced density maps showing the location of grapes in the images.  
The denisty map models required training examples that had the location of grapes manually labelled.  
Our current approach comes close to this level of accuracy using images that were taken 6 days prior to harvest without the the use of a calibration marker or manually marked-up images. 
This suggests that the proposed MTL model has great potential for use in the vineyard and more generally in agricultural settings.  




\begin{scriptsize}
\bibliography{estimage,thesisbib2}            
\bibliographystyle{iclr2020_conference}
\end{scriptsize}

\end{document}